%% file: main.tex
\documentclass[10pt]{article} 
\usepackage[preprint]{tmlr}

\input{math_commands.tex}

\usepackage{hyperref}
\usepackage{url}
\usepackage[american]{babel}

\usepackage{natbib} 
\usepackage{mathtools} 
\usepackage{amsfonts}
\usepackage{amsmath}
\usepackage{algorithm}
\usepackage{algpseudocode}
\usepackage{multirow}
\usepackage{cancel}
\usepackage[caption=false]{subfig}
\usepackage{booktabs} 
\usepackage{caption} 
\usepackage{tikz} 
\let\cite\citep

\title{Joint Causal Structure and Cluster Discovery Using Variational Inference}

\author{\name Avni Rajpal \email ai21resch04001@iith.ac.in \\
      \addr Department of Artificial Intelligence,
      Indian Institute of Technology Hyderabad, India
      \AND
      \name Anubhav Kumar \email ai24mtech12004@iith.ac.in \\
      \addr Department of Artificial Intelligence,
      Indian Institute of Technology Hyderabad, India
      \AND
      \name Rishabh Karnad \email ai22mtech12001@iith.ac.in \\
      \addr Department of Artificial Intelligence,
      Indian Institute of Technology Hyderabad, India
      \AND
      \name Mohammad Emtiyaz Khan \email emtiyaz.khan@riken.jp \\
      \addr RIKEN Center for AI Project, Tokyo, Japan
      \AND
      \name P.K. Srijith \email srijith@iith.ac.in \\
      \addr Department of Artificial Intelligence,
      Indian Institute of Technology Hyderabad, India}

\def\month{MM}  
\def\year{YYYY} 
\def\openreview{\url{https://openreview.net/forum?id=XXXX}} 

\begin{document}

\maketitle

\begin{abstract}
Causal discovery aims to understand the relationships between individual random variables. In many applications, such as brain imaging and climate modeling, it is more meaningful to consider interactions among groups of variables. Existing methods assume that knowledge of such groups or clusters is explicitly available when modeling interactions. However, in practice, these clusters as well as the causal relationships among them, are latent.  In this paper, we present a novel approach based on variational inference to simultaneously infer both the latent clusters and causal structures. We learn an approximate posterior over clusters and graph-structure by considering variational distributions based on  categorical and Bernoulli models respectively. We derive variational lower bounds and estimation techniques to learn variational and model   parameters. The effectiveness of our proposed methods for cluster and causal discovery are demonstrated on both synthetic and real data sets.
\end{abstract}

\section{Introduction}\label{sec:intro}
The ability to go beyond statistical associations and infer causal effect relationships between variables in a data-generating system is crucial in many scientific applications \citep{pearl2009causality,spirtes2000causation}. Causal discovery aims to learn causal structure over a set of random variables from observational data, often represented as a directed acyclic graph (DAG). 
Typically, causal learning methods assume that the data generating process can be modeled using a structural causal model (SCM) \citep{peters2017elements}. SCMs consist of a set of equations that represent independent mechanisms each of which produces scalar values and assumes that exogenous noise variables are independent. However, there are situations where it is more suitable to model mechanisms that generate vector-valued outputs.  For instance,  learning causal dependencies between brain regions rather than between individual neurons is more informative for neuroscientists \citep{semedo2020statistical,perich2020rethinking}. 

In literature,  DAGs over clusters, groups or partitions of variables are known as Group DAGs \citep{parviainen2017learning} or Cluster DAGs (C-DAGs) \citep{anand2023causal}. \citet{wahl2023vector} discuss how to do causal inference over a pair of clusters of variables when the clustering is known. Another instance where grouping of variables is useful is in the presence of latent confounding, where the assumption of independent noise is violated. One of the works that group variables in the presence of confounding between exogenous variables is  \citet{kawahara2010grouplingam}, which explicitly assumes a linear non-Gaussian model. Vector-valued SCMs (vSCMs) have been proposed recently by \citet{wahl2023foundations},  which generalizes classical SCMs to accommodate grouping assumptions defined by C-DAG. Most works on C-DAGs assume variable grouping is known through domain knowledge. Recently, \citet{niu2022learning} proposed a score-based approach to learn both clusters and DAGs jointly from the data, particularly when the underlying distribution is not faithful to the full DAG.  

In this paper, we address the problem of jointly inferring causal structure and clusters from the observational data.  We consider linear and non-linear Gaussian additive noise models based on vSCMs and  propose a novel approach that treats both the clusters and structures as latent variables. The proposed approach learns an approximate  posterior distribution over them by using variational inference~\cite{blei2017variational}. The proposed Bayesian treatment of learning distribution over clusters and structures can benefit in uncertainty quantification and active interventions. Bayesian learning has been found useful in the standard SCM setup in   \citet{annadani2021variational,lorch2021dibs,deleu2022bayesian}. However, developing a variational inference technique to infer both clusters and structures in vSCMs is challenging and we propose appropriate variational distributions and training methods to address this. The experimental results on synthetic and real data sets demonstrate the effectiveness of the proposed approach to jointly discover cluster and causal structure over clusters.    
\section{Background}\label{sec:background}
\subsection{Problem Statement}
\begin{figure}[t!]
\centering
\includegraphics[width=0.7\linewidth]{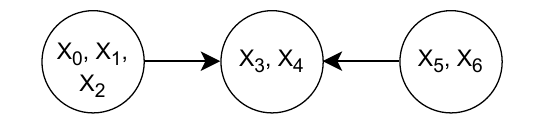}
\caption{Example of a cluster-DAG}
\label{fig:cdag-example}
\end{figure}

Let $\mathbf{X}$ be a $d$-dimensional random vector with elements $X_i \in \mathcal{R}$. Let $D$ be the dataset with $N$ observations of $\mathbf{X}$. Let $I$ be the index set with indices $\{1, \ldots, d\}$. We define a partitioning of the indices  as $C$ = $(C_1...C_k)$, denoting the $k$ clusters, where each cluster is a disjoint subset. Let $\mathbf{X}_{C_i}$ denote a vector containing all the values of $\mathbf{X}$ associated with index $C_i$. Let $k_i$ denote the number of elements in $\mathbf{X}_{C_i}$. A cluster-DAG or C-DAG is defined as the directed acyclic graph over the clusters in $C$, with an  adjacency matrix  denoted by $E_C$. An example of a C-DAG is given in Figure \ref{fig:cdag-example}.  We  consider the problem of cluster DAG discovery where we need to infer the clusters and connections in  the C-DAG $G_C=(C,E_C)$  from the observational samples $D$. 

\subsection{Vector-Valued Structural Causal Models}\label{sec:vscm}
 A vector-valued structural causal model (vSCM) \cite{wahl2023foundations} is a set of vector-valued assignments defined over these clusters:
\begin{align}
    \mathbf{X}_{C_i} := f_{C_i}(\text{Pa}(\mathbf{X}_{C_i}), \epsilon_i)
    \label{vscm}
\end{align}
where $\text{Pa}(\mathbf{X}_{C_i})$ is a vector containing a subset of $\mathbf{X}$ 
forming the  parents of $\mathbf{X}_{C_i}$ and $\epsilon_i \in \mathbb{R}^{k_i}$ is a noise vector whose elements are possibly dependent. 
We denote the number of elements in $\text{Pa}(\mathbf{X}_{C_i})$ as $\bar{k}_i$. The function $f_{C_i}: \mathbb{R}^{\bar{k}_i} \rightarrow \mathbb{R}^{k_i}$ is therefore a vector-valued function that defines the functional relationship between the vector $\mathbf{X}_{C_i}$ and its parent variables. The vSCM entails a graphical structure, which is a cluster DAG.  The vSCM induces a joint distribution over the observed variables  $p(\mathbf{X} | C, E_C; \Theta)$, where $\Theta$ are parameters of the distribution, that factorizes according to $C$ and $E_C$ as follows:
\begin{align}
p\left(\mathbf{X}|C, E_C;\Theta\right)=\prod_{i=1}^{k}p\left(\mathbf{X}_{C_i} | \text{Pa}(\mathbf{X}_{C_i}); \Theta_{C_i}\right) 
\end{align}
where $\Theta_{C_i}$ are the local parameters of the conditional distribution over $\mathbf{X}_{C_i}$, obtained from the global parameter $\Theta$ using index $C_i$.


\subsubsection{Linear Gaussian Vector Valued SCMs}\label{sec:lingauss}
A linear-Gaussian vSCM considers the following form between variables and their parents:
\begin{align}
    \mathbf{X}_{C_i} = \Theta_{C_i}^T\text{Pa}(\mathbf{X}_{C_i})  + \epsilon_i
    \label{eq:linGSCM}
\end{align}
where $\Theta_{C_i}$ is a matrix representing a linear mapping from $\text{Pa}(\mathbf{X}_{C_i})$ to $\mathbf{X}_{C_i}$ and $\epsilon_i$ has a multivariate Gaussian distribution with zero mean and covariance $\Sigma_{C_i}$. The distribution of $\mathbf{X}$ therefore decomposes into the product of Gaussians, 
\begin{align}\label{eqn:lingauss-likelihood}
P\left(\mathbf{X}|E_C,C,\Theta\right)=\prod_{i=1}^{k}\mathcal{N}\left(\mathbf{X}_{C_i} | \text{Pa}(\mathbf{X}_{C_i}); \mu_{C_i} ,\Sigma_{C_i}\right)  
\end{align}
where $\mu_{C_i} = \Theta_{C_i}^TPa(\mathbf{X}_{C_i})$, and $\Theta$ is the collection of $\Theta_{C_1}...\Theta_{C_k}$.
\subsubsection{Non-Linear Additive Noise Model}
We also consider the non-linear additive noise model, 
where we model the $f_{C_i}$ in \eqref{vscm} as a neural network with $\Theta_{C_i}$ as weights associated to the mapping between Pa$(\mathbf{X}_{C_i})$ and $\mathbf{X}_{C_i}$. We model all $f_{C_i}$ with a single fully connected neural network with parameters $\Theta$ by masking the inputs and outputs based on $C_i$ and $Pa(C_i)$\cite{lorch2021dibs,zheng2020learning}.
Therefore the distribution of $\mathbf{X}$ can be written as:
\begin{align}\label{eqn:nonlingauss-likelihood}
P(\mathbf{X}|E_C, C; \Theta) = \prod_{i=1}^k \mathcal{N}(\mathbf{X}_{C_i} | \mu_{C_i}, \Sigma_{C_i})
\end{align}
where $\mu_{C_i} = NN(Pa(X_{C_i}); \Theta_{C_i})$ is modelled as a neural network (NN).

\subsection{Marginal Likelihood of DAG}

For the case of standard DAG $G$, where the causal mechanism is linear Gaussian, the marginal likelihood   $P(D | G)$  can be computed in closed form~\cite{geiger2013parameter} and is given as 
\begin{align}
    P(D | G) = \prod_{i=1}^d \frac{P\Bigl(D^{\left(X_i, \text{Pa}_G\left(X_i\right)\right)}\Bigr)}{P\Bigl(D^{\text{Pa}_G\left(X_i\right)}\Bigr)}
\label{eqn:bge-factor}
\end{align}
where $\text{Pa}_{G}(X_i)$ are the parents of the node $X_i$ in DAG $G$. Consider $\mathbf{Y}$ to be any subset of $\mathbf{X}$, i.e., $\left(X_i, \text{Pa}_G\left(X_i\right)\right)$ or $\left(\text{Pa}_G\left(X_i\right)\right)$, then $D^{\mathbf{Y}}$ is the dataset $D$ restricted to observations of $\mathbf{Y}$.
For Gaussian DAG models with mean $\mu$ and precision matrix $W$ \citet{geiger2013parameter} considered Normal-Wishart distribution as parameter priors. Then the  parameters of Gaussian DAG model are marginalized out to give closed form expression for  $D^{\mathbf{Y}}$ as given in \citep{geiger2013parameter}:
$W \sim \mathcal{W}(T^{-1}, \alpha_w)$ and 
$\mu|W \sim \mathcal{N}(\nu, \alpha_\mu W)$, where $\nu$ is the prior mean, $T$ is the scale matrix, $\alpha_w > d-1$ is the degrees of freedom of the Wishart distribution and $\alpha_\mu > 0$ is a scaling parameter. 
\begin{equation}
\begin{split}
    p({D^{\mathbf{Y}}}) = \frac{1}{\pi^{lN / 2}}& \left( \frac{\alpha_\mu}{N + \alpha_\mu} \right)^{l / 2} \frac{\Gamma_l \left( (N + \alpha_w - d + l) / 2 \right)}{\Gamma_l \left( (\alpha_w - d + l) / 2 \right)}  \frac{|T_{\mathbf{YY}}|^{(\alpha_w - d + l) / 2}}{|R_{\mathbf{YY}}|^{(N + \alpha_w - d + l) / 2}}\\\\
    R &= T + S_N + \frac{N\alpha_\mu}{(N + \alpha_\mu)}(\nu - \bar{\mathbf{X}})(\nu - \bar{\mathbf{X}})^T
\end{split}
\label{eqn:bgelocal}
\end{equation}
where  $l$ is the dimension of $\mathbf{Y}$.  $\Gamma_l(.)$ is the multivariate Gamma function of dimension $l$.  $T_{\mathbf{YY}}$ and $R_{\mathbf{YY}}$ are  the sub-matrices of $T$ and $R$ respectively, containing the rows and columns corresponding to the elements in $\mathbf{Y}$.  $\bar{\mathbf{X}}$ and $S_N$ are the empirical mean and $(N-1) \times (N-1)$ empirical covariance of $\mathbf{X}$ respectively.



\begin{figure}[t!]
\centering
\includegraphics[width=0.22\linewidth]{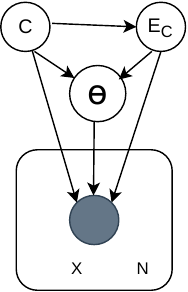}
\caption{Graphical model of the generative process}
\label{fig:graphical-model}
\end{figure}

\section{Methodology}
We propose a novel methodology to address the problem of jointly discovering clusters and their corresponding causal structures.  We provide a principled approach based on Bayesian learning and inference to jointly infer them. In particular, we rely on variational inference to learn an approximate posterior distribution over the latent clusters and structures from the observational data. 
In a finite sample setting, several C-DAGs can be consistent with the observed data and learning a distribution over the cluster DAGs (C-DAGs) help us to discover all of them. It also allows us to quantify uncertainty over C-DAGs which can be crucial for several high-risk applications. Further, it allows one to do  active interventions to obtain a better understanding of the underlying  causal model.

\subsection {Generative Model}
To formalize our approach, we begin by specifying the generative process underlying the observed data. The graphical model that defines the generative process is given in Figure \ref{fig:graphical-model}.  We assume that the samples of $\mathbf{X}$ in $D$ are generated by a linear Gaussian vSCM as in \eqref{eq:linGSCM} or non-linear additive model as in \eqref{eqn:nonlingauss-likelihood}.  In addition, $(C,E_C)$ are discrete \textit{latent} random variables, where $C$ specifies the cluster assignments that form the nodes in the C-DAG  and $E_C$  its edges. To maintain acyclicity, we assume $E_C$ to be a  strictly upper triangular matrix that represents the adjacency matrix over those cluster nodes. Then the joint distribution $P\left(C,E_C,D,\Theta\right)$ is given by:
\begin{align}
P\left(C,E_C,D,\Theta\right)\nonumber=P\left(D|C,E_C,\Theta\right)P\left(\Theta|C,E_C\right)P\left(E_C|C\right)P\left(C\right)
\end{align}
where  $P\left(C\right),P\left(E_C|C\right), P\left(\Theta|C,E_C\right)$ are the prior distributions and $P\left(D|\Theta, C,E_C\right)$ is the likelihood of modeling  data $D$ given model parameters $\Theta$, and latent $C$, and $E_C$. We would like to infer $C$, and $E_C$ by learning the posterior distribution over them following Bayesian principles. 
\subsection{Prior Distributions}
The Bayesian framework allows us to naturally encode domain knowledge about structures through priors. In addition, it plays a key role in estimating posterior over structures using Bayesian inference.
\paragraph{Prior over clusters} In our work, we assume a uniform prior over clusters.  With $d$ variables and $k$ clusters, the total number of possible cluster assignments is $k^{d}$.  The prior distribution is given by $P\left(C\right)=\frac{1}{k^d}$.
\paragraph{Prior over graph structure} 
 In the space of DAGs, the number of denser graphs is greater than the number of sparser graphs \citep{eggeling2019structure}. Consequently, fully connected graphs have a higher probability mass. To balance out this effect, we choose a prior over $E_C$,  which enforces sparsity and is given by:
\begin{align}
P\left( E_C |C \right)\propto \exp\left(-\lambda_s\left\| E_C \right\|_1 \right)
\label{eqn:graphprior}
\end{align}
where $\lambda_{s}$ is the tunable hyperparameter. Unlike causal discovery algorithms \citep{annadani2021variational,lorch2021dibs}, where the introduction of the acyclicity constraint in prior is essential to restrict the search space to DAGs, our parameterization of $E_C$ as an upper triangular matrix naturally enforces this constraint. 

We discuss the  posterior computation over $C$ and $E_C$ using Bayesian inference, for the linear and non-linear cases, separately in the following sections. 

\subsection{Linear Gaussian Vector Valued SCMs}
\subsubsection{Marginal Likelihood}\label{sec:bge}
If the data is generated using linear Gaussian vSCM, the distribution of $\mathbf{X}$ is multivariate Gaussian. Thus, the choice of prior over parameters $\Theta$ and the assumptions such as likelihood modularity, prior modularity, and global parameter independence used to derive the marginal likelihood of DAG ($p(D|G)$) in \ref{eqn:bge-factor} holds in this case. Consequently, the marginal likelihood $P(D|C,E_C)$ can be computed in closed form by integrating out $\Theta$.  Since, for C-DAGs, the DAG is over the clusters, we obtain marginal likelihood as:
\begin{align}
    P(D | C, E_C) = \prod_{i=1}^k \frac{P \Bigl(D^{\mathbf{X}_{C_i}, \text{Pa}\left(\mathbf{X}_{C_i}\right)}\Bigr)}{P\Bigl( D^{\text{Pa}\left(\mathbf{X}_{C_i}\right)}\Bigr)}
    \label{bge-factor-cdag}
\end{align}
where $\text{Pa}(X_{C_i})$ are the parents of the vector $X_{C_i}$ in C-DAG $(C,E_C)$. The expression for each of  the factors is obtained by using \eqref{eqn:bgelocal}.


\subsubsection{Bayesian Inference over C-DAGs}
By combining the prior and likelihood through Bayes rule, the posterior distribution over clusters and structures in  C-DAGs  is given as

\begin{align}
\nonumber
P\left( C,E_C|D\right)=&\frac{P\left( D|C,E_C\right)P\left( C \right)
P\left(E_C | C \right)}{P\left( D\right)} \\ \nonumber
=&\frac{P\left( D|C,E_C\right)P\left( C \right) P\left(E_C|C\right)}{\sum_{E_C}\sum_{C}P\left( D|C,E_C\right)P\left( C \right) P\left(E_C|C\right)} \\
\label{eqn:jointfactorization}
\end{align}

where $P\left(D\right)$ is the model evidence. 
Computing the posterior distribution in \eqref{eqn:jointfactorization} in closed form requires us to compute summations over all C-DAGs in the evidence term.  Since, the space of $C$ and $E_C$ over the $d$ variables and $k$ clusters grows in the order $\mathcal{O}\left(k^d\right)$ and $\mathcal{O}\left(k^2\right)$, respectively, this makes the computation infeasible. Hence, we will approximate the posterior $P\left(E_C,C|D\right)$ using variational distribution over $C$ and $E_C$ following the variational inference technique. 

\subsubsection{Variational Inference for C-DAGs}
Our goal is to learn the posterior distribution $P(C,E_C|D)$ given  the observed dataset $D$. The variational inference framework casts the problem of inference as an optimization problem where the true posterior is approximated using the tractable family of distributions $Q_\phi\left(C,E_C\right)$ whose parameters are learned by minimizing the KL divergence between the approximate posterior and the true posterior. We consider the following factorization of the variational distribution over $C$ and $E_C$, parameterized by $\mathbf{T}$ and $\mathbf{Z}$.  
\begin{align}
   P\left(C,E_C|D\right) \approx  Q_\phi\left(C,E_C|D\right) = Q_{\mathrm{\mathbf{T}}}\left( C\right)Q_{\mathbf{Z}}\left( E_C\right)
\label{eqn:approxvi1}
\end{align}
A direct minimization of the KL divergence requires computing the exact posterior. Variational Inference~\citep{blei2017variational} provides an elegant way to compute the variational distribution by maximizing the \textit{Evidence Lower Bound $\left(ELBO\right)$} instead, as in the following proposition.
\paragraph{Proposition 1.} Let $Q_{\mathrm{\mathbf{T}}}\left( C\right)$ and $Q_{\mathrm{\mathbf{Z}}}\left( E_C\right)$ be the variational distribution of $C$ and $E_C$ respectively. Then the \textit{evidence lower bound $\left(ELBO\right)$} is given by:
\begin{align*}
&\log P\left(D\right)\geq \mathcal{L}\left(\mathbf{Z,T}\right) \\
&=\mathbb{E}_{Q_{\mathbf{Z}}\left(E_C\right)}\mathbb{E}_{Q_{\mathbf{T}}\left(C\right)}\left[\log \frac{P\left(D|E_C,C\right)P\left(E_C,C\right)}{Q_{\mathbf{Z}}\left(E_C\right)Q_{\mathbf{T}}\left(C\right)}\right]
\end{align*}
\subsection{Non-linear Vector Valued SCM}

\subsubsection{Marginal Likelihood}
For linear vSCMs,  we obtained the marginal likelihood  by marginalizing out $\Theta$ parameters of the causal model: 
\begin{align}
P\left(D\right|C,E_C)=\int_{\Theta}P\left( D|\Theta,C,E_C\right)P\left(\Theta|C,E_C\right)d\theta
\label{eq:marginallikelihood-cdag}
\end{align}
However,  such a marginalization is intractable, and not computable in closed form in the case of non-Linear vSCMs. So instead, we consider the following joint distribution and perform a point estimation of $\Theta$:
\begin{align}
P\left(C,E_C,D|\Theta\right)\nonumber=P\left(D|C,E_C;\Theta\right)P\left(E_C|C\right)P\left(C\right)
\end{align}
\subsubsection{Variational Inference for Non-linear vSCM}
We can write the posterior $P(C,E_C|D;\Theta)$ as:
\begin{align}
\nonumber
P\left( C,E_C|D;\Theta\right)=\frac{P\left( D|C,E_C;\Theta\right)P\left( C,E_C \right)}{P\left( D;\Theta\right)} \\ \nonumber
\end{align}
Due to the intractability in computing the marginal $P(D|\Theta)$, we resort to variational inference to obtain the posterior approximation. We consider the same variational approximation as in \eqref{eqn:approxvi1}. 
However, the expression of the Evidence Lower Bound (ELBO), is different in this case and depends on the model parameters $\Theta$. 
\paragraph{Proposition 2.} Let $Q_{\mathrm{\mathbf{T}}}\left( C\right)$ and $Q_{\mathrm{\mathbf{Z}}}\left( E_C\right)$ be the variational distribution of $C$ and $E_C$ respectively, and $\Theta$ be the parameters of the neural network. Then the \textit{evidence lower bound $\left(ELBO\right)$} is given by:
\begin{align*}
&\log P\left(D;\Theta\right)\geq \mathcal{L}\left(\mathbf{Z,T},\Theta\right) \\
&=\mathbb{E}_{Q_{\mathbf{Z}}\left(E_C\right)}\mathbb{E}_{Q_{\mathbf{T}}\left(C\right)}\left[\log \frac{P\left(D|E_C,C;\Theta\right)P\left(E_C,C\right)}{Q_{\mathbf{Z}}\left(E_C\right)Q_{\mathbf{T}}\left(C\right)}\right]
\end{align*}

\subsection{Variational Families}

Our goal is to approximate an intractable joint distribution with a variational family that is tractable as well as flexible to model complex distributions over C-DAGs. We will first discuss the choice of variational family for $C$, then discuss the choice for $E_C$.

\subsubsection{Distribution over Clusters}  In our work, we considered two models for $C$ a) Factorized model and b) Linear autoregressive model. 

In $\textbf{Factorized model}$, we assume that each of the dimensions in $C$ is independent and the variational posterior of $C$ can be written as the product of independent categorical random variables given by:
\begin{align}
\label{eqn:varclustfactor}
\nonumber
Q_{\mathrm{\mathbf{T}}}\left(C\right) &= \prod_{m=1}^{d}Q_{\mathrm{\mathbf{t_m}}}\left(C_{m}\right)\\
Q_{\mathrm{\mathbf{t_m}}}\left(C_{m}=i\right) &=\frac{\exp\left(t_{mi}\right)}{\sum_{n=1}^{k}\exp\left(t_{mn}\right)}\\
\nonumber
\end{align}
where $i=\{1, \ldots, k\}, \mathrm{\mathbf{T}} \in \mathbb{R}^{d\times k}$ is a matrix of logits.

The factorized model is a simpler model; such an approach may not be sufficient to capture the correlated density. There is inherent dependence among the dimensions that allows them to group together, which motivates us to look for a more expressive distribution. One natural way to capture this dependence is to learn the autoregressive distribution on $C$. 

In $\textbf{linear autoregressive model}$, we model the variational posterior of $C$ as a product of conditionals, and each conditional is modeled with a categorical distribution given by:
\begin{align}
Q_{\mathrm{\mathbf{T}}}\left(C\right) &= \prod_{m=1}^{d}Q_{\mathrm{\mathbf{t_m}}}\left(C_{m}|C_{1:m-1}\right)\\
Q_{\mathrm{\mathbf{t_m}}}\left(C_{m}=i|C_{1:m-1}\right) &=\frac{\exp\left(t_{mi}\right)}{\sum_{n=1}^{k}\exp\left(t_{mn}\right)}\\
\nonumber
\end{align}
where $i=\{1, \ldots, k\}, \mathrm{\mathbf{T}} \in \mathbb{R}^{d\times k}$ is a matrix of logits and $\mathbf{t_{m}}$ are the parameters of each conditional that are estimated using the linear autoregressive model. 
The linear autoregressive model is given by the following equation: 
\begin{align}
\begin{split}
h_m&=Ah_{m-1}+b_{hh}+BC_{m-1}+b_{ch} \\
t_m&=Dh_m+c_{ho}
\end{split}
\label{eq:linearar}
\end{align}
where $C_{m-1}, h_m ,t_m$ are the input, hidden state, and logits for $m^{th}$ variable. $A,B,D$ and $b$'s are parameters of the model. 
\subsubsection{Distribution over graphs}  In our work, we considered four models for the adjacency matrix $E_C$: a) Independent binary model, b) Linear autoregressive model, c) Conditional model, and d) Autoregressive conditional model. Since $E_C$ is upper triangular, the dimension of $E_C$ is $k\times(k-1)/2$. 

In the $\textbf{independent binary model}$, we assume that each entry in $E_C$ is an independent binary random variable. The variational distribution over $E_C$ is given as:
\begin{align}
\nonumber
Q_\mathbf{Z}\left(E_C\right)=\prod_{i=1}^{k\times(k-1)/2}Q_{z_i}\left(e_{c_i}\right) = \prod_{i=1}^{k\times(k-1)/2}\frac{1}{1+\exp\left( -z_{i}\right)}\\
\label{eqn:vargraph}
\end{align}
where $\mathrm{\mathbf{Z}}\in\mathbb{R}^{k\times(k-1)/2}$ is the vector of logits.
The factorized model assumes independent edges, which may not be sufficient to capture the structural dependencies within the graph. To address this, we introduce three more expressive variational families for $E_C$.

In $\textbf{linear autoregressive model}$ for graphs, we capture the topological dependence among the edges by modeling the variational posterior of $E_C$ as a product of conditionals, given by:
\begin{align}
Q_\mathbf{Z}\left(E_C\right) &= \prod_{i=1}^{k\times(k-1)/2}Q_{z_i}\left(e_{c_i}|e_{c_{1:i-1}}\right)\\
Q_{z_i}\left(e_{c_i}=1|e_{c_{1:i-1}}\right) &=\frac{1}{1+\exp\left(-z_{i}\right)}
\end{align}
where $z_i$ are the logits estimated using the linear autoregressive model:
\begin{align}
\begin{split}
h_i&=Ah_{i-1}+b_h+Be_{c_{i-1}} \\
z_i&=Dh_i+c_{he}
\end{split}
\label{eq:linearar_graph}
\end{align}
where $e_{c_{i-1}}, h_i, z_i$ are the input, hidden state, and logit for the $i^{th}$ edge. $A, B, D, b_h,$ and $c_{he}$ are parameters of the model.

However, the variational families discussed earlier fail to accommodate the dependence of graph ($E_C$) on the cluster assignments ($C$). In the $\textbf{conditional model}$, we explicitly model the dependence of the edges on the cluster assignments $C$:
\begin{align}
Q_\mathbf{Z}\left(E_C|C\right) &= \prod_{i=1}^{k\times(k-1)/2}Q_{z_i}\left(e_{c_i}|C\right)
\end{align}
Here, the vector of logits $\mathrm{\mathbf{Z}}\in\mathbb{R}^{k\times(k-1)/2}$ is defined as a linear transformation of the flattened one-hot representation of the cluster assignments, denoted as $C_{\text{flat}} \in \{0,1\}^{dk}$:
\begin{align}
\mathrm{\mathbf{Z}} &= W_{E_C}C_{\text{flat}} + b_{E_C}\label{eq:conditional_graph}
\end{align}
where $W_{E_C} \in \mathbb{R}^{k(k-1)/2 \times dk}$ and $b_{E_C}$ are the learned weight matrix and bias parameters, respectively.

Finally, to simultaneously capture both the internal graph topology and its dependence on the global cluster assignments, we combine these approaches into an $\textbf{autoregressive conditional model}$. The variational posterior integrates both the previous edge states and the cluster assignments:
\begin{align}
Q_\mathbf{Z}\left(E_C|C\right) &= \prod_{i=1}^{k\times(k-1)/2}Q_{z_i}\left(e_{c_i}|e_{c_{1:i-1}}, C\right)
\end{align}
The logits $z_i$ are computed by incorporating both the autoregressive hidden state $h_i$ and the global cluster context $C_{\text{flat}}$:
\begin{align}
\begin{split}
h_i&=Ah_{i-1}+b_h+Be_{c_{i-1}} \\
z_i&=Dh_i+W_{C}C_{\text{flat}}+c_{bias}
\end{split}
\label{eq:ar_conditional_graph}
\end{align}
where $W_{C}$ projects the global cluster representation into the logit space for the $i^{th}$ edge alongside the autoregressive representation.
%
\begin{algorithm}[t]
  \caption{C-DAG Learning Using VI}
  \label{algo:CDAG-VI}
  \begin{algorithmic}[1]
    \State \textbf{Input:} Dataset D
    \State Initialize parameters $\mathbf{Z},\mathbf{T}$
    \While{not converged}
        \State $(C^i,E_C^i)_{i=1}^{M} \sim Q_{\mathbf{T}}\left(C\right), Q_{\mathbf{Z}}\left(E_C\right)$
        \State  $G_\mathbf{Z}=\nabla_\mathbf{Z}\mathcal{L}\left(\mathbf{Z},\mathbf{T}\right)$  with $(C^i,E_C^i)_{i=1}^{M}$ \Comment{use eq(\ref{eqn:gradzelbo})}
        
        \State  $G_\mathbf{T}=\nabla_\mathbf{T}\mathcal{L}\left(\mathbf{Z},\mathbf{T}\right)$  with $(C^i,E_C^i)_{i=1}^{M}$ \Comment{use eq(\ref{eqn:gradTelbo})}
        
        \State $\mathbf{Z} \leftarrow $ Update parameters using $G_\mathbf{Z}$
        \State $\mathbf{T} \leftarrow $ Update parameters using $G_\mathbf{T}$
    \EndWhile
  \end{algorithmic}
\end{algorithm}
\subsection{Gradient Approximation of ELBO}
\label{subsec:gradients}
As described previously, in order to approximate the posterior over $(C,E_C)$ using variational inference, we estimate the parameters of the 
 variational posterior by maximizing \textit{ELBO}. We optimize \textit{ELBO} using a gradient-based approach. This requires us to estimate the gradients $\nabla_{(\mathbf{Z},\mathbf{T})}\mathcal{L}\left(\mathbf{Z,T}\right)$. Since $C,E_C$ are discrete variables, we approximate the gradients using the score function estimator, REINFORCE~\citep{williams1992simple}. REINFORCE estimators are known to suffer from variance issues\citep{lievin2020optimal}, and in our work, we used an exponential moving average as the baseline $b$ similar to that used in \citet{annadani2021variational} for variance reduction. Following this, the gradients are computed as:
\begin{equation}
\begin{split}
\nabla_\mathbf{Z}\mathcal{L}\left(\mathbf{Z,T}\right)=&\mathbb{E}_{Q_{\mathbf{Z}}\left(E_C\right)}\mathbb{E}_{Q_{\mathbf{T}}\left(C\right)}\Bigl[\nabla_\mathbf{Z}\log Q_{\mathbf{Z}}\left(E_C\right)\\&\times\Bigl(\log \frac{P\left(D|E_C,C\right)P\left(E_C,C\right)}{Q_{\mathbf{Z}}\left(E_C\right)Q_{\mathbf{T}}\left(C\right)}-b\Bigr)\Bigr]
\end{split}
\label{eqn:gradzelbo}
\end{equation}
\begin{equation}
\begin{split}
\nabla_\mathbf{T}\mathcal{L}\left(\mathbf{Z,T}\right)=&\mathbb{E}_{Q_{\mathbf{Z}}\left(E_C\right)}\mathbb{E}_{Q_{\mathbf{T}}\left(C\right)}\Bigl[\nabla_\mathbf{T}\log Q_{\mathbf{T}}\left(C\right)\\&\times\Bigl(\log \frac{P\left(D|E_C,C\right)P\left(E_C,C\right)}{Q_{\mathbf{Z}}\left(E_C\right)Q_{\mathbf{T}}\left(C\right)}-b\Bigr)\Bigr]
\end{split}
\label{eqn:gradTelbo}
\end{equation}
The expectations in the gradient computations are approximated using Monte Carlo sampling.  A summary of the proposed methodology is given in Algorithm \ref{algo:CDAG-VI}. 


For the non-linear formulation, we explore several variational families: for the clusters $C$, we utilize the factorized model \eqref{eqn:varclustfactor} and the linear autoregressive model \eqref{eq:linearar}; for the graph $E_C$, we consider the independent binary model \eqref{eqn:vargraph}, the linear autoregressive model \eqref{eq:linearar_graph}, the conditional model \eqref{eq:conditional_graph}, and the autoregressive conditional model \eqref{eq:ar_conditional_graph}. Following Proposition 2, the gradients for variational parameters are computed as:
\begin{equation}
\begin{split}
\nabla_\mathbf{Z}\mathcal{L}\left(\mathbf{Z},\mathbf{T},\Theta\right)=&\mathbb{E}_{Q_{\mathbf{Z}}\left(E_C\right)}\mathbb{E}_{Q_{\mathbf{T}}\left(C\right)}\Bigl[\nabla_\mathbf{Z}\log Q_{\mathbf{Z}}\left(E_C\right)\\\times\Bigl(&\log \frac{P\left(D|E_C,C;\Theta\right)P\left(E_C,C\right)}{Q_{\mathbf{Z}}\left(E_C\right)Q_{\mathbf{T}}\left(C\right)}-b\Bigr)\Bigr]
\end{split}
\label{eqn:gradzelbonl}
\end{equation}
\begin{equation}
\begin{split}
\nabla_\mathbf{T}\mathcal{L}\left(\mathbf{Z},\mathbf{T},\Theta\right)=&\mathbb{E}_{Q_{\mathbf{Z}}\left(E_C\right)}\mathbb{E}_{Q_{\mathbf{T}}\left(C\right)}\Bigl[\nabla_\mathbf{T}\log Q_{\mathbf{T}}\left(C\right)\\\times\Bigl(&\log \frac{P\left(D|E_C,C;\Theta\right)P\left(E_C,C\right)}{Q_{\mathbf{Z}}\left(E_C\right)Q_{\mathbf{T}}\left(C\right)}-b\Bigr)\Bigr]
\end{split}
\label{eqn:gradTelbonl}
\end{equation}

As the distributions on $C$ and $E_C$ are independent of $\Theta$ in the ELBO, the gradient for $\Theta$ is computed as:
\begin{align}
    \nabla_{\Theta}&\mathcal{L}\left(\mathbf{Z},\mathbf{T},\Theta\right) \nonumber \\&= \mathbb{E}_{Q_{\mathbf{Z}}\left(E_C\right)}\mathbb{E}_{Q_{\mathbf{T}}\left(C\right)}\left[\nabla_\Theta\log P\left(D|E_C,C;\Theta\right)\right]
\label{eqn:gradtheta}
\end{align}
All of the expectations are approximated with Monte Carlo sampling, the proposed methodology is summarized in Algorithm \ref{algo:CDAG-nn-VI}.

\begin{algorithm}[t]
  \caption{Non-Linear C-DAG Learning}
  \label{algo:CDAG-nn-VI}
  \begin{algorithmic}[1]
    \State \textbf{Input:} Dataset D
    \State Initialize parameters $\mathbf{Z},\mathbf{T}, \Theta$
    \While{not converged}
        \State $(C^i,E_C^i)_{i=1}^{M} \sim Q_{\mathbf{T}}\left(C\right), Q_{\mathbf{Z}}\left(E_C\right)$
        \State  $G_\mathbf{Z}=\nabla_\mathbf{Z}\mathcal{L}\left(\mathbf{Z},\mathbf{T},\Theta\right)$  with $(C^i,E_C^i)_{i=1}^{M}$ \Comment{use eq(\ref{eqn:gradzelbonl})}
        
        \State  $G_\mathbf{T}=\nabla_\mathbf{T}\mathcal{L}\left(\mathbf{Z},\mathbf{T},\Theta\right)$  with $(C^i,E_C^i)_{i=1}^{M}$ \Comment{use eq(\ref{eqn:gradTelbonl})}

        \State $G_\mathbf{\Theta} = \nabla_\mathbf{\Theta}\mathcal{L}\left(\mathbf{Z},\mathbf{T},\Theta\right)$ with $(C^i,E_C^i)_{i=1}^{M}$ \Comment{use eq(\ref{eqn:gradtheta})}
        
        \State $\mathbf{Z} \leftarrow $ Update parameters using $G_\mathbf{Z}$
        \State $\mathbf{T} \leftarrow $ Update parameters using $G_\mathbf{T}$
        \State $\mathbf{\Theta} \leftarrow $ Update parameters using $G_\mathbf{\Theta}$
    \EndWhile
  \end{algorithmic}
\end{algorithm}

\section{Experimental Results}
In this section, we study the empirical performance of our method compared to the baseline \citep{niu2022learning} on the synthetic and real world datasets. Although most of the literature focuses on causal discovery for scalar variables, there are several works that study causal interaction on a group of variables \citep{wahl2023foundations,anand2023causal}. 

However, these works study theoretical aspects \citep{chalupka2016multi,rubenstein2017causal} or the works in the direction of causal discovery assume that information about the groups is known \citep{wahl2023vector,parviainen2017learning}. To the  best of our knowledge, \citet{niu2022learning} is the only work in the literature that focuses on the simultaneous estimation of clusters and graphs from observational data. This work gives the point estimates of the learned C-DAGs, however, we use bootstrapping to obtain multiple estimates in order to compare the performance on multiple metrics. We use this as the baseline in our experiments. 
We will first discuss the evaluation criteria, then we will discuss the datasets considered in our work, the experimental setup and their corresponding results.
\subsection{Evaluation Metrics}
In case of smaller dimensions, i.e., $d<=4$, enumeration of all DAGs is feasible and the exact posterior can be computed. In such scenarios, performance evaluation is done by computing the distance between the variational posterior and the ground truth posterior. Typically, \textbf{Hellinger distance (HD)} is used as a distance metric between distributions \citep{annadani2021variational}. Compared to the space of all DAGs, the space of C-DAGs grows more slowly, with its complexity primarily determined by the number of clusters. Example for $k=4$, one can enumerate C-DAGs upto $d \leq 7$.
The other metrics used to evaluate the performance of causal discovery algorithms over DAGs, specifically for higher dimensions,  are the structural Hamming distance (SHD) and the area under receiver operating curve (AUROC) \citep{annadani2021variational,lorch2021dibs}. These metrics are intended to capture the similarity between the ground truth and the estimated graph structure. We take inspiration from the causal discovery literature and use the following metrics for evaluation.

\paragraph{Expected Expanded-Graph SHD (ExG-SHD)}: Structural Hamming distance \cite{zheng2018dags} measures how many edges are different between two DAG structures. The \textbf{Expected SHD [E-SHD]} over the posterior is utilized in literature \citep{annadani2021variational,lorch2021dibs} to evaluate the performance of Bayesian causal discovery methods. We also compute the same metric on $E_C$ to evaluate the accuracy of the variational posterior.

\begin{figure}[!t]
    \includegraphics[width=0.9\linewidth]{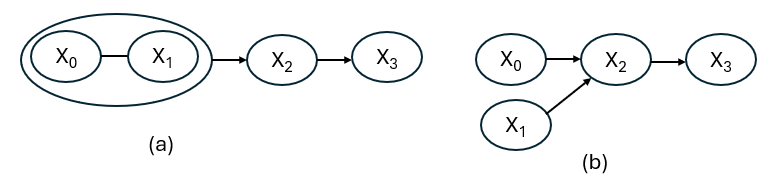}
    \caption{Example of a) C-DAG b) Expanded Graph}
    \label{fig:expandedgraph}
\end{figure}

However, this metric doesn't take into account the approximation error due to the variational posterior of $C$. In order to capture the impact of approximation error of both $C,E_C$ on the joint posterior, we define a new object called the \textbf{expanded-graph} and compute expected structural hamming distance on it. We call this metric \textbf{Expected Expanded Graph SHD (ExG-SHD)}. An example of a C-DAG and its corresponding expanded graph is shown in Figure \ref{fig:expandedgraph}. We define the expanded-graph as $G_{\text{expand}} = CE_CC^T$. and the ExG-SHD with respect to the ground truth expanded-graph $G^*$ is given by:
\begin{align}
\begin{split}
\mathbb{E}_{Q_{\textbf{T}}(C)Q_{\textbf{Z}}(E_C)}\left[\text{SHD}\right]&\approx \frac{1}{N}\sum_{i=1}^{N}\Bigl[\text{SHD}(G^{(i)}_{\text{expand}},G^*)\Bigr] \\
C^{(i)},E_C^{(i)} &\sim Q_{\textbf{T}}(C),Q_{\textbf{Z}}(E_C) \\ 
G^{(i)}_{\text{expand}}&=C^{(i)}E_C^{(i)}C^{(i)T}
\label{eq:shd}
\end{split}
\end{align}

\paragraph{Area Under Receiver Operating Curve (AUROC)}: Another commonly used metric in causal discovery literature \citep{annadani2021variational,lorch2021dibs} to evaluate the performance of Bayesian models is AUROC. We extend the metric to C-DAGs, where we compute edge beliefs over expanded graphs and compute the Receiver Operating Curve (ROC). The area under this curve is used as a metric to evaluate the performance.

\paragraph{Expected Rand Index (E[RI])}: The similarity between the clusterings of the C-DAGs obtained from the estimated posterior are measured using rand-index \citep{hubert1985comparing}. The expected rand-index with respect to the ground truth clustering $C^*$ is given by:
\begin{align}
\begin{split}
\mathbb{E}_{Q_{\textbf{T}}(C)}\left[\text{RI}\right]&\approx \frac{1}{N}\sum_{i=1}^{N}\Bigl[\text{RI}(C^{(i)},C^*)\Bigr] \\
C^{(i)} &\sim Q_{\textbf{T}}(C)
\end{split}
\end{align}
\begin{table}[t]
\centering
\caption{Results on linear synthetic datasets. ExG-SHD, E-SHD, and HD: lower is better; E-RI and AUC: higher is better. NA: Not Applicable.}
\label{tab:linear_synthetic}

\subfloat[4 variables, 3 clusters\label{tab:4var3clus}]{%
\resizebox{0.48\linewidth}{!}{%
\begin{tabular}{|c|c|c|c|c|c|c|}
\hline
\begin{tabular}[c]{@{}c@{}}Data-\\ sets\end{tabular} &
  Methods &
  \begin{tabular}[c]{@{}c@{}}ExG-\\ SHD\end{tabular} &
  \begin{tabular}[c]{@{}c@{}}E-\\ RI\end{tabular} &
  AUC &
  \begin{tabular}[c]{@{}c@{}}E-\\ SHD\end{tabular} &
  HD \\ \hline
\multirow{4}{*}{\begin{tabular}[c]{@{}c@{}}4var\\ 3clus\\ fork\end{tabular}} &
  Baseline &
  3.6 &
  0.55 &
  0.55 &
  1.85 &
  NA \\ \cline{2-7} 
 &
  Fact-C-E &
  3.99 &
  0.66 &
  0.74 &
  0.99 &
  1.0 \\ \cline{2-7} 
 &
  AR-C-Fact-E &
  \textbf{0.02} &
  0.99 &
  \textbf{1.0} &
  \textbf{0.0} &
  \textbf{0.7} \\ \cline{2-7} 
 &
  Fact-C-AR-E &
  2.0 &
  \textbf{1.0} &
  0.55 &
  2.0 &
  0.7 \\ \hline
\multirow{4}{*}{\begin{tabular}[c]{@{}c@{}}4var\\ 3clus\\ chain\end{tabular}} &
  Baseline &
  3.6 &
  0.55 &
  0.42 &
  1.9 &
  NA \\ \cline{2-7} 
 &
  Fact-C-E &
  1.0 &
  0.66 &
  \textbf{0.79} &
  0.0 &
  1.0 \\ \cline{2-7} 
 &
  AR-C-Fact-E &
  \textbf{0.03} &
  0.67 &
  0.6 &
  \textbf{0.0} &
  \textbf{0.71} \\ \cline{2-7} 
 &
  Fact-C-AR-E &
  2.00 &
  \textbf{1.00} &
  0.58 &
  2.0 &
  \textbf{0.71} \\ \hline
\multirow{4}{*}{\begin{tabular}[c]{@{}c@{}}4var\\ 3clus\\ vstruc\end{tabular}} &
  Baseline &
  2.2 &
  0.8 &
  0.83 &
  1.0 &
  NA \\ \cline{2-7} 
 &
  Fact-C-E &
  1.0 &
  1.0 &
  1.0 &
  0.95 &
  0.97 \\ \cline{2-7} 
 &
  AR-C-Fact-E &
  \textbf{0.0} &
  \textbf{1.0} &
  \textbf{1.0} &
  \textbf{0.0} &
  \textbf{0.56} \\ \cline{2-7} 
 &
  Fact-C-AR-E &
  3.19 &
  0.66 &
  0.66 &
  1.0 &
  0.89 \\ \hline
\end{tabular}
}
}
\hfill
\subfloat[7 variables, 3 clusters\label{tab:7var3clus}]{%
\resizebox{0.48\linewidth}{!}{%
\begin{tabular}{|c|c|c|c|c|c|c|}
\hline
\begin{tabular}[c]{@{}c@{}}Data-\\ sets\end{tabular} &
  Methods &
  \begin{tabular}[c]{@{}c@{}}ExG-\\ SHD\end{tabular} &
  \begin{tabular}[c]{@{}c@{}}E-\\ RI\end{tabular} &
  AUC &
  \begin{tabular}[c]{@{}c@{}}E-\\ SHD\end{tabular} &
  HD \\ \hline
\multirow{4}{*}{\begin{tabular}[c]{@{}c@{}}7var\\ 3clus\\ fork\end{tabular}} &
  Baseline &
  10.8 &
  0.3 &
  0.55 &
  1.9 &
  NA \\ \cline{2-7} 
 &
  Fact-C-E &
  10.0 &
  1.0 &
  \textbf{0.77} &
  2.0 &
  0.71 \\ \cline{2-7} 
 &
  AR-C-Fact-E &
  \textbf{4.0} &
  \textbf{1.0} &
  0.74 &
  2.0 &
  \textbf{0.71} \\ \cline{2-7} 
 &
  Fact-C-AR-E &
  13.9 &
  0.65 &
  \textbf{0.81} &
  \textbf{1.0} &
  1.0 \\ \hline
\multirow{4}{*}{\begin{tabular}[c]{@{}c@{}}7var\\ 3clus\\ chain\end{tabular}} &
  Baseline &
  10.1 &
  0.6 &
  0.47 &
  0.21 &
  NA \\ \cline{2-7} 
 &
  Fact-C-E &
  14.9 &
  0.71 &
  0.56 &
  0.99 &
  1.0 \\ \cline{2-7} 
 &
  AR-C-Fact-E &
  10.0 &
  0.61 &
  \textbf{0.72} &
  \textbf{0.0} &
  \textbf{0.99} \\ \cline{2-7} 
 &
  Fact-C-AR-E &
  \textbf{5.0} &
  \textbf{0.8} &
  \textbf{0.72} &
  2.0 &
  1.0 \\ \hline
\multirow{4}{*}{\begin{tabular}[c]{@{}c@{}}7var\\ 3clus\\ vstruc\end{tabular}} &
  Baseline &
  10.95 &
  0.2 &
  0.68 &
  1.2 &
  NA \\ \cline{2-7} 
 &
  Fact-C-E &
  11.9 &
  0.71 &
  0.8 &
  1.0 &
  1.0 \\ \cline{2-7} 
 &
  AR-C-Fact-E &
  \textbf{6.8} &
  \textbf{0.75} &
  \textbf{0.95} &
  \textbf{1.0} &
  \textbf{0.99} \\ \cline{2-7} 
 &
  Fact-C-AR-E &
  10.3 &
  0.63 &
  0.75 &
  \textbf{1.0} &
  \textbf{0.99} \\ \hline
\end{tabular}
}
}
\end{table}
\subsection{Experimental Setup}
In our work, we conducted experiments on both synthetic and real-world datasets to evaluate how close our algorithm is able to approximate the posterior over $C,E_C$. In our experiments, we approximated $P\left(C|D\right)$ with both our linear and non-linear approaches and compared the results with respect to the baseline~\cite{niu2022learning} for all datasets.  Moreover, since the baseline is a non-Bayesian approach, to compare it with our Bayesian approach, we use bootstrapping \cite{DAGBootstrap} to obtain multiple estimates by running the baseline on $20$ bootstrapped datasets. For each run, we used $1000$ iterations.

For both factorized and linear autoregressive models, we performed a hyperparameter tuning search for different learning rates $(0.1,0.01,0.001)$, optimizers (SGD, RMSprop, Adam) and learning rate schedulers (exponential, Cosine-Annealed LR). The RMSprop optimizer with a learning rate of $0.01$ annealed by the exponential scheduler with gamma $0.9$ at every $500$ epoch; these settings were found to be optimal across all datasets. For the linear autoregressive model, we searched over the dimensions $(4,8,16,32,48,64)$ of the hidden state \eqref{eq:linearar}. We chose the hyperparameters that result in maximum  \textit{ELBO}. Maximization of \textit{ELBO}  with respect to variational parameters is known to suffer from local optima, thus the solution heavily depends on the initial state. In our work, for each dataset and for each of the models, we do 10 random restarts. We run each of the experiments for $10,000$ epochs, with early stopping.

The Neural model was trained with the AdamW optimizer, with a cosine decay learning rate scheduler going from $5\times10^{-3}$ to $2\times10^{-5}$ for the first half of training. The 2 layer Neural Network hidden states were searched between $(16, 32, 64, 128)$. For the real world datasets, we do $75,000$ epochs, while for synthetic datasets we do $50,000$ epochs of training, and for all datasets and models we do 10 random restarts.


We consider multiple variants of the proposed model, differing in their variational families. Fact-C-E employs factorized models for both $C$ and $E_C$, while AR-C-Fact-E introduces an autoregressive model for $C$ alongside a factorized $E_C$. The remaining variants all maintain a factorized structure for $C$: Fact-C-AR-E applies an autoregressive model to $E_C$, Fact-C-Cond-E uses a conditional distribution for $E_C\vert{}C$, and Fact-C-AR-Cond-E leverages an autoregressive conditional architecture for $E_C\vert{}C$.

\subsection{Synthetic Dataset}
We generate samples for synthetic data using linear Gaussian vector-valued SCMs given by \eqref{eqn:lingauss-likelihood}.  
Based on the number of clusters and fixed group sizes, we randomly generate the cluster assignments. To obtain the adjacency matrix over the clusters, we sample Erdos-Renyi graphs with edge probability $0.4$ and ensure that the adjacency matrix generated is strictly upper triangular. 
Each of the elements of $\Theta \in \mathbb{R}^{d\times d}$ is sampled uniformly in $\{-2.,-5.\} \cup \{2.,5.\}$. The edge weights $\Theta_{C_i}$ are then obtained by masking $\Theta$ with the expanded graph. For the generation of synthetic data, we used a specific form of noise covariance $\Sigma_{C_i}$ that is added to each cluster. $\Sigma_{C_i}$ is given by:
\begin{align*}
    \Sigma_{C_i} = \sigma \left( I_{k_i} + \rho\left(F_{k_i \times k_i}-I_{k_i}\right) \right)
\end{align*}
where $I_{k_i} \in \mathbb{R}^{k_i \times k_i}$ is an identity matrix and $F_{k_i \times k_i} \in \mathbb{R}^{k_i \times k_i}$ is  matrix of  all 1's. $\rho$ allows control over the correlation between variables, and $\sigma$ adjusts the spread of the distribution. For all conditional distributions in \eqref{eqn:lingauss-likelihood}, we used a covariance with the same $\sigma$ and $\rho$. Such a choice of sigma ensures that clusters in each group are correlated. For our experiments, we generated $N=1000$ samples for $d=\{4,7\}$, $k=3$ with $\sigma=0.1$ and $\rho=0.9$. The fixed group sizes considered are $(2,1,1)$ and $(3,2,2)$ for $d=4$ and $d=7$ respectively. For 3 nodes, the interesting cases are chains, forks and v-structures, hence, we consider only those to generate our data. The C-DAGs considered are provided in the supplementary material. The non-linear dataset is generated in a similar manner, however a random non-linearity is applied to $\Theta_{C_i}^T$Pa$(\mathbf{X}_{C_i})$ to generate $\mathbf{X}_{C_i}$. We generate $N=3000$ samples for the non-linear data.

\begin{table}[t]
\centering
\caption{Results on nonlinear synthetic datasets. ExG-SHD and E-SHD: lower is better; E-RI and AUC: higher is better.}
\label{tab:nonlin_synthetic}

\subfloat[4 variables, 3 clusters\label{tab:nonlin4var}]{%
\resizebox{0.48\linewidth}{!}{%
\begin{tabular}{|c|c|c|c|c|c|}
\hline
\begin{tabular}[c]{@{}c@{}}Data-\\ sets\end{tabular} &
  Methods &
  \begin{tabular}[c]{@{}c@{}}ExG-\\ SHD\end{tabular} &
  \begin{tabular}[c]{@{}c@{}}E-\\ RI\end{tabular} &
  AUC &
  \begin{tabular}[c]{@{}c@{}}E-\\ SHD\end{tabular} \\ \hline
\multirow{6}{*}{\begin{tabular}[c]{@{}c@{}}4var\\fork\end{tabular}}
& Baseline
& 2.90 & \textbf{0.70} & 0.63 & 2.00 \\ \cline{2-6}
& Fact-C-E
& \textbf{0.00} & 0.67 & \textbf{1.00} & \textbf{1.00} \\ \cline{2-6}
& AR-C-Fact-E
& 4.88 & 0.59 & 0.87 & \textbf{1.00} \\ \cline{2-6}
& Fact-C-AR-E
& \textbf{0.00} & 0.67 & \textbf{1.00} & 2.00 \\ \cline{2-6}
& Fact-C-Cond-E
& \textbf{0.00} & 0.67 & \textbf{1.00} & \textbf{1.00} \\ \cline{2-6}
& Fact-C-AR-Cond-E
& \textbf{0.00} & 0.67 & \textbf{1.00} & 1.99 \\ \hline

\multirow{6}{*}{\begin{tabular}[c]{@{}c@{}}4var\\chain\end{tabular}}
& Baseline
& 2.80 & 0.70 & 0.62 & 2.00 \\ \cline{2-6}
& Fact-C-E
& 0.01 & \textbf{1.00} & \textbf{1.00} & \textbf{0.00} \\ \cline{2-6}
& AR-C-Fact-E
& 4.77 & 0.57 & 0.87 & 1.00 \\ \cline{2-6}
& Fact-C-AR-E
& 2.00 & \textbf{1.00} & 0.92 & 1.00\\ \cline{2-6}
& Fact-C-Cond-E
& \textbf{0.00} & \textbf{1.00} & \textbf{1.00} & \textbf{0.00} \\ \cline{2-6}
& Fact-C-AR-Cond-E
& 2.00 & 0.99 & 0.87 & 1.00 \\ \hline

\multirow{6}{*}{\begin{tabular}[c]{@{}c@{}}4var\\vstruc\end{tabular}}
& Baseline
& 5.00 & 0.60 & 0.50 & 2.40 \\ \cline{2-6}
& Fact-C-E
& 1.00 & 0.67 & \textbf{1.00} & 1.01 \\ \cline{2-6}
& AR-C-Fact-E
& 5.27 & 0.61 & 0.94 & \textbf{1.00} \\ \cline{2-6}
& Fact-C-AR-E
& \textbf{0.01} & \textbf{0.83} & \textbf{1.00} & \textbf{1.00}\\ \cline{2-6}
& Fact-C-Cond-E
& 1.00 & 0.67 & \textbf{1.00} & \textbf{1.00} \\ \cline{2-6}
& Fact-C-AR-Cond-E
& 1.00 & 0.67 & 0.97 & \textbf{1.00} \\ \hline
\end{tabular}
}
}
\hfill
\subfloat[7 variables, 3 clusters\label{tab:nonlin7var}]{%
\resizebox{0.48\linewidth}{!}{%
\begin{tabular}{|c|c|c|c|c|c|}
\hline
\begin{tabular}[c]{@{}c@{}}Data-\\ sets\end{tabular} &
  Methods &
  \begin{tabular}[c]{@{}c@{}}ExG-\\ SHD\end{tabular} &
  \begin{tabular}[c]{@{}c@{}}E-\\ RI\end{tabular} &
  AUC &
  \begin{tabular}[c]{@{}c@{}}E-\\ SHD\end{tabular} \\ \hline
\multirow{6}{*}{\begin{tabular}[c]{@{}c@{}}7var\\fork\end{tabular}}
& Baseline
& 12.00 & 0.20 & 0.58 & 1.80 \\ \cline{2-6}
& Fact-C-E
& 6.00 & \textbf{0.81} & 0.99 & 1.00 \\ \cline{2-6}
& AR-C-Fact-E
& 17.05 & 0.57 & 0.48 & 1.00 \\ \cline{2-6}
& Fact-C-AR-E
& \textbf{0.01} & 0.71 & \textbf{1.00} & 1.68 \\ \cline{2-6}
& Fact-C-Cond-E
& \textbf{0.01} & 0.71 & \textbf{1.00} & \textbf{0.59} \\ \cline{2-6}
& Fact-C-AR-Cond-E
& \textbf{0.01} & 0.71 & \textbf{1.00} & 2.00 \\ \hline

\multirow{6}{*}{\begin{tabular}[c]{@{}c@{}}7var\\chain\end{tabular}}
& Baseline
& \textbf{11.70} & \textbf{0.80} & 0.74 & 1.90 \\ \cline{2-6}
& Fact-C-E
& 13.00 & 0.67 & 0.71 & \textbf{1.00} \\ \cline{2-6}
& AR-C-Fact-E
& 17.23 & 0.59 & 0.61 & \textbf{1.00} \\ \cline{2-6}
& Fact-C-AR-E
& 12.00 & 0.71 & 0.77 & \textbf{1.00} \\ \cline{2-6}
& Fact-C-Cond-E
& 12.00 & 0.71 & \textbf{0.78} & \textbf{1.00} \\ \cline{2-6}
& Fact-C-AR-Cond-E
& 12.00 & 0.71 & \textbf{0.78} & \textbf{1.00} \\ \hline

\multirow{6}{*}{\begin{tabular}[c]{@{}c@{}}7var\\vstruc\end{tabular}}
& Baseline
& 13.80 & 0.60 & 0.57 & 1.50 \\ \cline{2-6}
& Fact-C-E
& \textbf{0.00} & \textbf{0.81} & \textbf{1.00} & 0.57 \\ \cline{2-6}
& AR-C-Fact-E
& 17.81 & 0.58 & 0.59 & 1.00 \\ \cline{2-6}
& Fact-C-AR-E
& \textbf{0.00} & \textbf{0.81} & \textbf{1.00} & 0.07 \\ \cline{2-6}
& Fact-C-Cond-E
& \textbf{0.00} & \textbf{0.81} & \textbf{1.00} & \textbf{0.00} \\ \cline{2-6}
& Fact-C-AR-Cond-E
& \textbf{0.00} & \textbf{0.81} & \textbf{1.00} & \textbf{0.00} \\ \hline
\end{tabular}
}
}
\end{table}

Since, the number of variables is small, enumeration of all CDAGs is possible, hence besides computing the metrics defined in the previous section, we were able to compute the distance between exact posterior and estimated posterior using Hellinger distance for the linear models. Baseline being a non-Bayesian approach, Hellinger distance computation is not applicable to the baseline. 

The results for the synthetic datasets are provided in Tables \ref{tab:4var3clus}, \ref{tab:7var3clus}, \ref{tab:nonlin4var} and \ref{tab:nonlin7var}. Across all tested metrics on both the 4-variable and 7-variable synthetic datasets, we can see that our approach consistently performs better than the baseline method. In the case of linear formulations, we note that the autoregressive cluster model generally outperforms the other variants, indicating the effectiveness of capturing the dependencies amongst the clusters. The baseline method struggles to accurately reconstruct the graphs, which is evident in its higher ExG-SHD scores, the metric that measures full graph recovery. In the case of linear formulations, the autoregressive cluster model generally outperforms the simpler factorized variant, indicating the practical necessity of capturing the dependencies amongst the clusters. For instance, in the 4-variable fork dataset (Table \ref{tab:4var3clus}), AR-C-Fact-E achieves an ExG-SHD of just 0.02 and an E-RI score of 0.99, tightly matching the ground truth.

In the nonlinear formulations, the performance gains are even more pronounced. The neural network is effectively able to learn the non-linearities allowing for our approach to vastly outperform than the baseline in most cases. As seen in Table \ref{tab:nonlin4var}, several non-linear variants (such as Fact-C-E, Fact-C-AR-E, and Fact-C-Cond-E) achieve a flawless ExG-SHD of 0.00 for the 4-variable fork and v-structure configurations. Furthermore, we notice that both autoregressive and conditional models for edges outperform the factorized edge models, indicating that capturing topological edge dependencies yields better structural estimations than assuming the edges are entirely independent. Interestingly, the autoregressive cluster models do not perform as well in these settings. 

\subsection{Real World Datasets}
In this section, we discuss the performance of our algorithm compared to the baseline on the two real-world datasets namely, Protein dataset and Climate dataset. The details of each of these datasets are provided in the appendix.

\begin{table}[]
\caption{Results on real-world datasets. NA means Not Applicable. ExG-SHD, E-SHD, HD lower is better. E-RI and AUC higher is better.}
\label{tab:sachs}
\centering
\resizebox{0.6\textwidth}{!}{%
\begin{tabular}{|c|c|c|c|c|c|c|}
\hline
\begin{tabular}[c]{@{}c@{}}Data-\\ sets\end{tabular} &
  Methods &
  \begin{tabular}[c]{@{}c@{}}ExG-\\ SHD\end{tabular} &
  \begin{tabular}[c]{@{}c@{}}E-\\ RI\end{tabular} &
  AUC &
  \begin{tabular}[c]{@{}c@{}}E-\\ SHD\end{tabular} &
  HD \\ \hline
\multirow{6}{*}{\begin{tabular}[c]{@{}c@{}}Protein\\ Dataset\end{tabular}} &
  Baseline &
  3.0 &
  0.55 &
  1.0 &
  0.3 &
  NA \\ \cline{2-7} 
 &
  Lin Fact-C-E &
  27.4 &
  0.57 &
  1.0 &
  1.0 &
  1.0 \\ \cline{2-7} 
 &
  Lin AR-C-Fact-E &
  \textbf{0.0} &
  0.56 &
  \textbf{1.0} &
  \textbf{0.0} &
  \textbf{0.7} \\ \cline{2-7} 
 &
  Neur Fact-C-E &
  0.04 &
  0.74 &
  \textbf{1.0} &
  \textbf{0.0} &
  NA \\ \cline{2-7} 
 &
  Neur AR-C-Fact-E &
  \textbf{0.0} &
  0.56 &
  \textbf{1.0} &
  \textbf{0.0} &
  NA \\ \cline{2-7} 
 &
  Neur Fact-C-Cond-E &
  \textbf{0.0} &
  \textbf{0.80} &
  \textbf{1.0} &
  \textbf{0.0} &
  NA \\ \hline
\multirow{6}{*}{\begin{tabular}[c]{@{}c@{}}Climate\\ Dataset\end{tabular}} &
  Baseline &
  15.75 &
  0.55 &
  0.6 &
  1.1 &
  NA \\ \cline{2-7} 
 &
  Lin Fact-C-E &
  20.1 &
  0.55 &
  0.31 &
  0.0 &
  1.0 \\ \cline{2-7} 
 &
  Lin AR-C-Fact-E &
  16 &
  \textbf{1.0} &
  0.3 &
  0.99 &
  \textbf{0.54} \\ \cline{2-7} 
 &
  Neur Fact-C-E &
  0.08 &
  \textbf{1.0} &
  \textbf{1.0} &
  \textbf{0.0} &
  NA \\ \cline{2-7} 
 &
  Neur AR-C-Fact-E &
  16 &
  0.43 &
  0.5 &
  0.66 &
  NA \\ \cline{2-7} 
 &
  Neur Fact-C-Cond-E &
  \textbf{0.04} &
  \textbf{1.0} &
  \textbf{1.0} &
  \textbf{0.0} &
  NA \\ \hline
\end{tabular}}
\end{table}
The results for the real-world datasets are provided in Table \ref{tab:sachs}. Our approach performs better than the baseline in most of the metrics for real-world datasets. For both datasets, we were able to enumerate the exact posterior. The Hellinger distance indicates that the autoregressive model for clusters is able to approximate the true posterior better than the factorized cluster model.  

In the case of the protein dataset, the ground-truth C-DAG does not have an edge, consisting instead of two separate, unconnected clusters. From Table \ref{tab:sachs}, the metrics ExG-SHD and E-SHD that capture the precision of the structure are at the perfect values of 0 for the linear autoregressive cluster model, indicating its effectiveness on real-world datasets. In comparison, the baseline yields a much higher ExG-SHD of 3.0. The Neural models perform even better, with almost all of them achieving near perfect scores for ExG-SHD and E-SHD and demonstrating better clusterings with high E-RI scores. Neural Conditional Edge model in particular has the highest E-RI of 0.80, while sharing ExG-SHD and E-SHD scores of the linear autoregessive cluster model. This is in comparison to the baseline which yields an E-RI of 0.55.

The ground truth C-DAG for climate dataset has two clusters and an edge between. The neural models near perfectly recover the structure and clusters, indicating the non-linear nature of the underlying data. The baseline model struggles here, resulting in a high ExG-SHD of 15.75 and an E-SHD of 1.1. In stark contrast, the neural formulations almost perfectly recover both the cluster assignments and the network structure. The Neural Conditional Edge model achieves a remarkably low ExG-SHD score of 0.04 and an E-SHD score of 0.0, accompanied with a perfect AUC of 1.0. It also demonstrates perfect E-RI score of 1.0, compared to the baseline score of 0.55. This exceptional recovery strongly suggests that the underlying causal mechanisms are inherently non-linear, and conditional modeling has notable improvement over the other methods.


\section{Conclusion}
We developed a variational inference technique for learning  posterior over clusters and structures in  C-DAGs, considering Bernoulli distribution over edges and categorical distribution over cluster assignments. Additionally, we examined two approaches for modeling cluster assignments: a factorized model and an autoregressive linear model to capture dependencies among assignments. We also examined multiple approaches to model the graph structure over these clusters, including an independent binary model, a linear autoregressive model, a conditional model, and an autoregressive conditional model.

To accommodate more complex data-generating processes, we successfully extended our framework to non-linear vector-valued SCMs by parameterizing the causal mechanisms with neural networks. The inference techniques were developed  for both the linear and non linear models of vSCMs. The experiments on synthetic and real data sets showed the effectiveness of our approaches compared to the baselines.  In the future, we would like to extend the work to scale it for high-dimensional datasets.



 
\bibliography{main}

\newpage

\onecolumn

\title{Joint Causal Structure and Cluster Discovery Using Variational Inference\\(Supplementary Material)}
\maketitle


\appendix

\section{Proof of Proposition 1}
\label{appendix:elboproof}

\begin{align}
\nonumber
&\arg\min_{\mathbf{Z},\mathbf{T}}KL(Q_\mathbf{Z}\left(E_C\right)Q_\mathbf{T}\left(C\right)||P\left(E_C,C|D\right)) \\ \nonumber
&=\mathbb{E}_{Q_{\mathbf{Z}}\left(E_C\right)}\mathbb{E}_{Q_{\mathbf{T}}\left(C\right)}\Bigl[\log Q_\mathbf{Z}\left(E_C\right) + \log Q_\mathbf{T}\left(C\right)-\log\left(\frac{P\left(D,E_C,C\right)}{P\left(D\right)}\right)\Bigr] \\ \nonumber
&=\mathbb{E}_{Q_{\mathbf{Z}}\left(E_C\right)}\mathbb{E}_{Q_{\mathbf{T}}\left(C\right)}\Bigl[\log Q_\mathbf{Z}\left(E_C\right)+\log Q_\mathbf{T}\left(C\right)-\log P\left(D,E_C,C\right)+\log P\left(D\right)\Bigr]
\end{align}
which gives us lower bound on the marginal log likelihood of the data
\begin{align}
\nonumber
\log P\left(D\right)&\geq\mathbb{E}_{Q_{\mathbf{Z}}\left(E_C\right)}\mathbb{E}_{Q_{\mathbf{T}}\left(C\right)}\Bigl[\log \frac{P\left(D|E_C,C\right)P\left(E_C,C\right)}{Q_{\mathbf{Z}}\left(E_C\right)Q_{\mathbf{T}}\left(C\right)}\Bigr]
\end{align}

\section{Proof of Proposition 2}
\label{appendix:elboproof2}
Consider the marginal log likelihood of the data 
\begin{align}
\nonumber
&\log P(D;\Theta) = \log\sum_{E_C}\sum_{C}P(D|C,E_C;\Theta)P\left(E_C|C\right)P\left(C\right) \\ \nonumber
& = \log\sum_{E_C}\sum_{C}\frac{P(D|C,E_C;\Theta)P\left(E_C|C\right)P\left(C\right)Q_{\mathbf{Z}}\left(E_C\right)Q_{\mathbf{T}}\left(C\right)}{Q_{\mathbf{Z}}\left(E_C\right)Q_{\mathbf{T}}\left(C\right)} \\ \nonumber
\end{align}
By definition of Expectation:
\begin{align}
\nonumber
&= \log \mathbb{E}_{Q_{\mathbf{Z}}\left(E_C\right)}\mathbb{E}_{Q_{\mathbf{T}}\left(C\right)}\Bigl[\frac{P(D|C,E_C;\Theta)P\left(E_C|C\right)P\left(C\right)}{Q_{\mathbf{Z}}\left(E_C\right)Q_{\mathbf{T}}\left(C\right)}\Bigr]\end{align}
By Jensen's inequality:
\begin{align}
\nonumber&\geq\mathbb{E}_{Q_{\mathbf{Z}}\left(E_C\right)}\mathbb{E}_{Q_{\mathbf{T}}\left(C\right)}\Bigl[\log \frac{P\left(D|E_C,C;\Theta\right)P\left(E_C,C\right)}{Q_{\mathbf{Z}}\left(E_C\right)Q_{\mathbf{T}}\left(C\right)}\Bigr]
\end{align}
\section{Data Generating C-DAGs for Real and Synthetic Datasets}
In our work, we showed the performance of our algorithm on both synthetic and real-world datasets. Synthetic datasets are generated assuming a linear vector structural causal model. Synthetic data sets of $d=4,7$ and $k=3$ were generated. For 3 nodes the interesting cases are chain, fork and v-structure, hence, we focused on these scenarios in our work. 

Finding real datasets where information about DAGs over clusters is known is very difficult. In our work, we considered 2 real-world datasets Protein dataset and climate dataset. Protein dataset does not have any explicit information about ground-truth C-DAG. Looking at the DAG which is generated by experts, we can explicitly see two clusters which we considered as the ground truth. In \citet{wahl2023vector} demonstrated their performance on the climate dataset. This motivates us to use this dataset. The ground truth CDAG used to generate data for each of the cases is given below:

\subsection{Data Generating C-DAG for 4 variable 3 cluster dataset}
\begin{figure}[!h]
\centering
\includegraphics[width=0.65\linewidth, keepaspectratio]{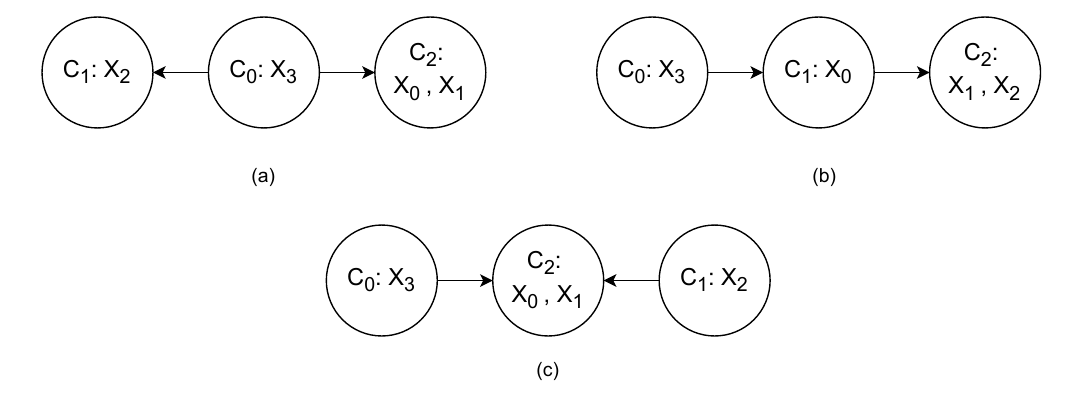}
\caption{Data Generating CDAG for 4 variable 3 cluster synthetic dataset a) fork b) chain c) v-structure}
\label{fig:4var3clus:gtcdag}
\end{figure}

The Figure(\ref{fig:4var3clus:gtcdag}) shows data generating C-DAGs for $d=4$ and $k=3$.

\subsection{Data Generating C-DAG for 7 variable 3 cluster dataset}
\begin{figure}[!h]
\centering
\includegraphics[width=0.8\linewidth, keepaspectratio]{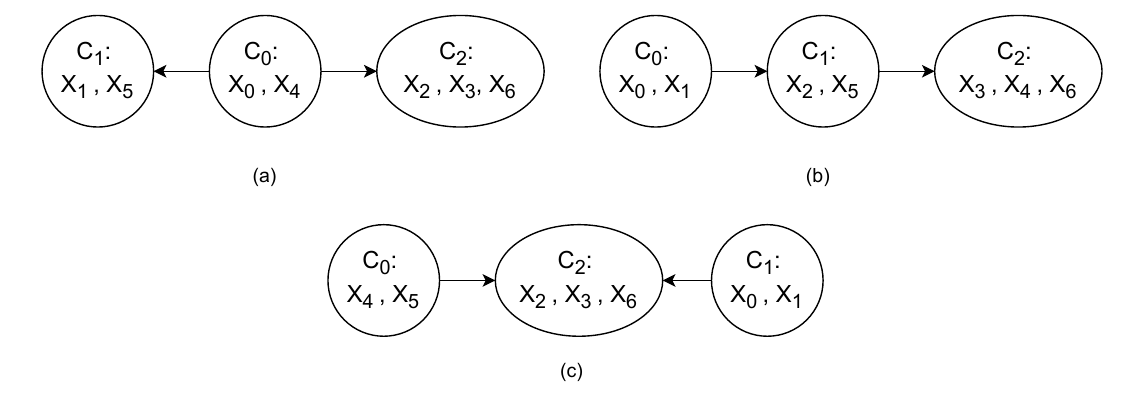}
\caption{Data Generating CDAG for 7 variable 3 cluster synthetic dataset a) fork b) chain c) v-structure}
\label{fig:7var3clus:gtcdag}
\end{figure}

The Figure(\ref{fig:7var3clus:gtcdag}) shows data generating C-DAGs for $d=7$ and $k=3$.

\subsection{Non-Linearities for synthetic Non-Linear data}

Our synthetic non-linear dataset is generated from the C-DAGs described in Figures \ref{fig:4var3clus:gtcdag} and \ref{fig:7var3clus:gtcdag}. An element-wise non-linear operation is applied after the linear combination of parents to ensure non-linearity. Table \ref{tab:listnonlin} describes the used non-linearities.

\begin{table}[!h]
\centering
\caption{Element-wise non-linearities applied after linear combination of parents}
\label{tab:listnonlin}
\begin{tabular}{|cc|c|c|}
\hline
\multicolumn{2}{|c|}{Dataset}        & C$_1$   & C$_2$ \\ \hline
\multicolumn{1}{|c|}{\multirow{3}{*}{\begin{tabular}[c]{@{}c@{}}4 variable \\ 3 cluster\end{tabular}}} & fork & Tanh   & Tanh \\ \cline{2-4} 
\multicolumn{1}{|c|}{} & chain       & Sin  & ReLU \\ \cline{2-4} 
\multicolumn{1}{|c|}{} & v-structure & -  & Cubic Polynomial \\ \hline
\multicolumn{1}{|c|}{\multirow{3}{*}{\begin{tabular}[c]{@{}c@{}}7 variable \\ 3 cluster\end{tabular}}} & fork & Sin & ReLU \\ \cline{2-4} 
\multicolumn{1}{|c|}{} & chain       & ReLU  & Sin  \\ \cline{2-4} 
\multicolumn{1}{|c|}{} & v-structure & -  & Tanh   \\ \hline
\end{tabular}
\end{table}

\subsection{Details for  Protein Dataset}
\begin{figure}[!h]
\centering
\subfloat[Data-Generating DAG\label{fig:sachs:gtdag}]{%
  \includegraphics[width=0.3\linewidth, keepaspectratio]{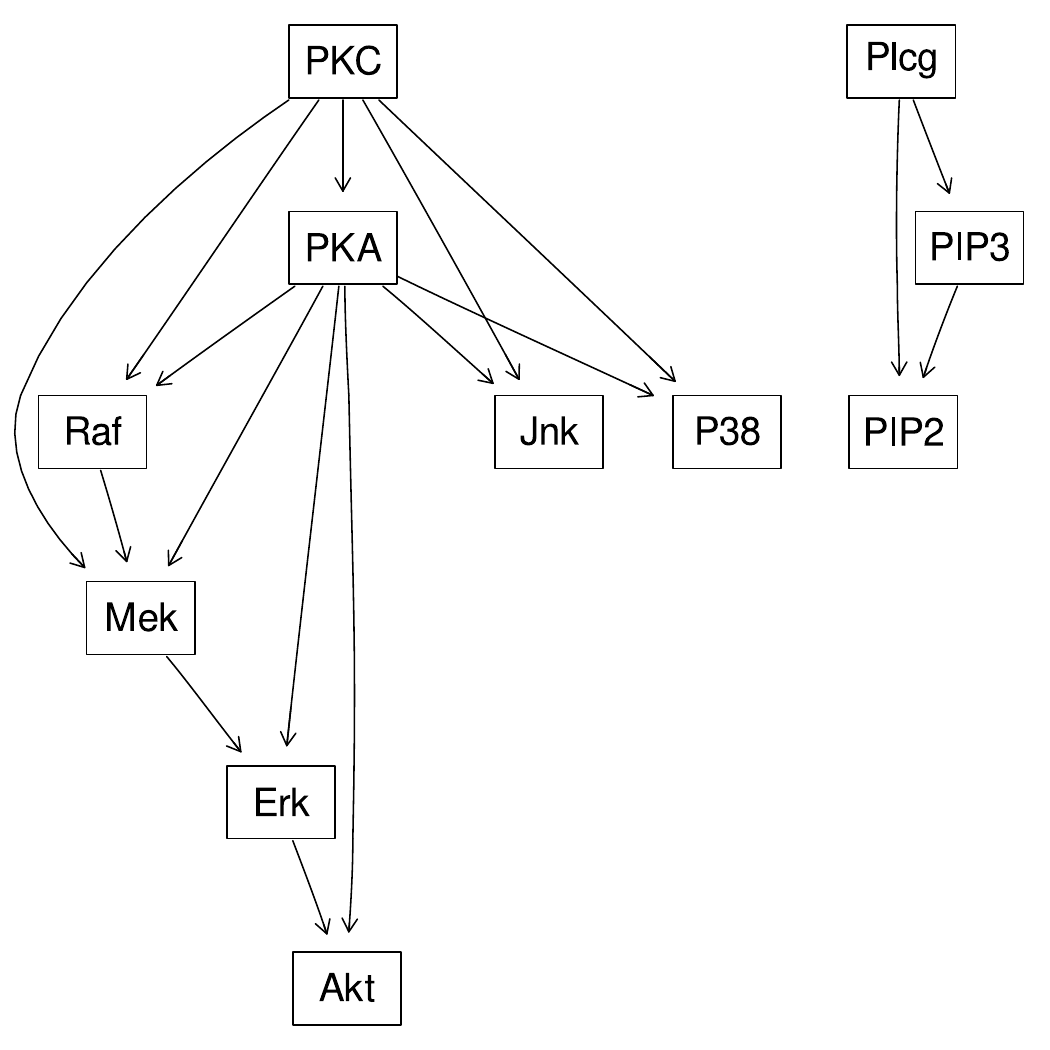}%
}\qquad
\subfloat[Ground-truth C-DAG\label{fig:sachs:gtcdag}]{%
  \includegraphics[width=0.3\linewidth, keepaspectratio]{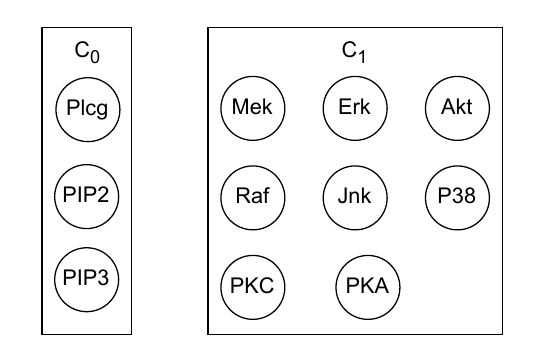}%
}
\caption{Sachs Dataset}
\end{figure}

We evaluated the performance of our algorithm on the protein signalling dataset \cite{sachs2005causal}, which contains $d=11$ features and $N=853$ observations, with a ground truth graph provided by experts shown in Figure \ref{fig:sachs:gtdag}. There is no true C-DAG known for this dataset. However, according to the ground truth structure, it is evident that there are two separate clusters which are not connected, i.e., one cluster containing nodes \{Plcg, PIP3, PIP2\} and the rest of the nodes in another cluster. 
The Figure(\ref{fig:sachs:gtdag}) shows data generating DAG given by experts. In our work we used  Figure(\ref{fig:sachs:gtcdag}) as the ground-truth  C-DAG and is used for reporting metrics.

\subsection{Details for Climate dataset}

We also evaluate our method on the climate dataset from \cite{wahl2023vector}. The dataset consists of samples of surface temperatures in two geographical regions - the Tropical Pacific (ENSO) and British Columbia (BCT). The causal influence of temperature variations in the ENSO region on the BCT region is established in climate science, and the method described by \cite{wahl2023vector} recovers this relationship with around 59\% accuracy. The dataset is generated by first deseasonalizing each sample by subtracting the monthly mean temperature of that geographical location, and Gaussian kernel smoothing with a bandwidth of $\sigma=120$ months is used to remove long-term trends. Next, the temperatures from the months of October, November and December are averaged for the ENSO region, while those for the months of January, February and March are averaged for BCT in order to obtain 73 samples of annual temperatures, with BCT offset by one time lag. The resulting dataset contains 2 clusters, whose individual dimensionalities can be controlled by adjusting the size of grid boxes within each region. 
The climate data set consists of surface temperature samples from two regions, ENSO and BCT. 

\begin{figure}[!h]
\centering
\includegraphics[width=0.3\linewidth, keepaspectratio]{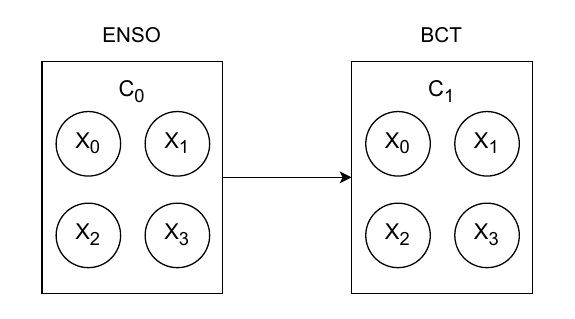}
\caption{Ground truth C-DAG for climate dataset}
\label{fig:enso:gtcdag}
\end{figure}

We considered surface temperature from four randomly chosen grid points in each region, hence $d=8$ and $k=4$. Figure(\ref{fig:enso:gtcdag}) shows the ground-truth  C-DAG for the climate dataset used to calculate the metrics.

\section{Experimental Details}
\begin{figure}[!t]
\centering
\includegraphics[width=0.5\linewidth, keepaspectratio]{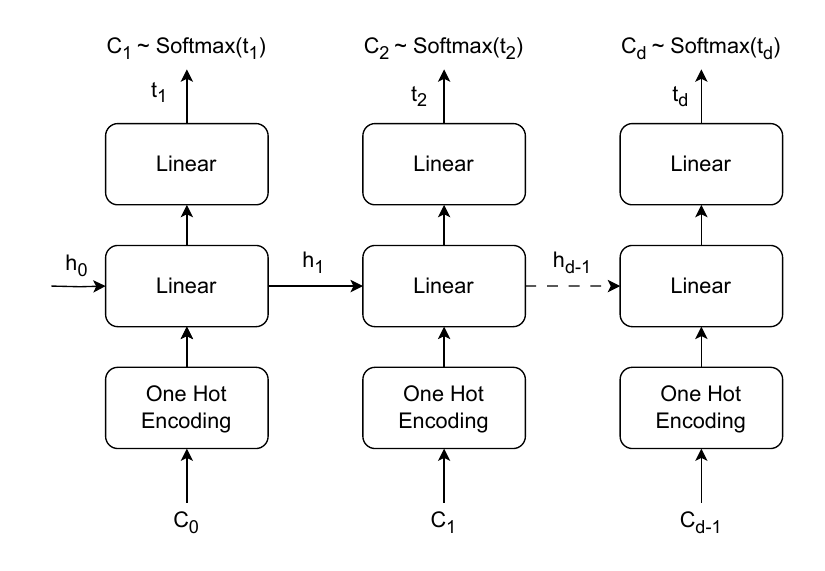} 
\caption{Autoregressive generation of cluster assignments using the linear model}
\label{fig:linearcat}
\end{figure}

For both factorized and linear autoregressive models eq(\ref{eq:linearar}),figure(\ref{fig:linearcat}), we performed a hyperparameter tuning search for different learning rates $(0.1,0.01,0.001)$, optimizers (SGD, RMSprop, Adam) and learning rate schedulers (exponential, Cosine-Annealed LR). The RMSprop optimizer with a learning rate of $0.01$ annealed by the exponential scheduler with gamma $0.9$ at every $500$ epoch; these settings were found to be optimal across all datasets. We also varied $\lambda_s=(1,10,100)$. For 7 variable 3 cluster chain dataset $\lambda_{s}=100$ was used, for remaining datasets $\lambda_{s}=1$ was found to be suitable. 

In our work, for each dataset and for each of the models, we do 10 random restarts. We run each of the experiments for $10,000$ epochs. We used Xavier uniform for parameter initialization. All experiments are run on Tesla V100-SXM2-32GB GPU. For the linear autoregressive model, we searched over the dimensions $(4,8,16,32,48,64,128)$ of the hidden state. The final hidden state dimension that gave the best results is provided in Table(\ref{tab:hidden-state}). In addition, how much variation is across multiple restarts is shown Table(\ref{tab:variance:linearmodel}). 

For our neural formulation, we use a 2 hidden layer MLP with input and output layers being $d$ dimensional, and hidden layer activations being softplus. The hidden layer dimension of 64 for both layers was found to be optimal among $(16,32,64,128)$. We use the AdamW optimizer to apply weight decay regularization for our neural parameters $\Theta$. $\lambda_s = 1000$ was found to be optimal for the protein signalling dataset, $\lambda_s = 30$ for the climate dataset, and $\lambda_s = 500$ across the synthetic nonlinear datasets. 

Each experiment was run $10$ times, and the learning was done for $75000$ epochs for the real world datasets and for $50000$ epochs for the synthetic datasets. A cosine decay scheduler for learning rate was used for the first half of training to lower the learning rate from $5\times10^{-3}$ to $2\times10^{-5}$. ELBO is approximated using $30$ samples, and the performance metrics are calculated using $1000$ samples from the predictive distribution.

\begin{table}[!h]
\centering
\caption{Linear Autoregressive Model hidden-state dimension choices}
\label{tab:hidden-state}
\begin{tabular}{|cc|c|}
\hline
\multicolumn{2}{|c|}{Dataset} &
  \begin{tabular}[c]{@{}c@{}}Hidden-State\\ Dimension\end{tabular} \\ \hline
\multicolumn{1}{|c|}{\multirow{3}{*}{\begin{tabular}[c]{@{}c@{}}4 variable \\ 3 cluster\end{tabular}}} &
  fork &
  32 \\ \cline{2-3} 
\multicolumn{1}{|c|}{} & chain       & 4   \\ \cline{2-3} 
\multicolumn{1}{|c|}{} & v-structure & 4   \\ \hline
\multicolumn{1}{|c|}{\multirow{3}{*}{\begin{tabular}[c]{@{}c@{}}7 variable \\ 3 cluster\end{tabular}}} &
  fork &
  8 \\ \cline{2-3} 
\multicolumn{1}{|c|}{} & chain       & 128 \\ \cline{2-3} 
\multicolumn{1}{|c|}{} & v-structure & 48  \\ \hline
\multicolumn{2}{|c|}{Sachs}          & 4   \\ \hline
\multicolumn{2}{|c|}{Climate}        & 4   \\ \hline
\end{tabular}
\end{table}

\begin{table}[!h]
\centering
\caption{Variation across random restarts for autoregressive linear cluster model for selected hidden dimension size}
\label{tab:variance:linearmodel}
\begin{tabular}{|cc|c|c|}
\hline
\multicolumn{2}{|c|}{Dataset}        & E-RI±std   & E-SHD±std \\ \hline
\multicolumn{1}{|c|}{\multirow{3}{*}{\begin{tabular}[c]{@{}c@{}}4 variable \\ 3 cluster\end{tabular}}} & fork & 0.7±0.1   & 1.08±0.51 \\ \cline{2-4} 
\multicolumn{1}{|c|}{} & chain       & 0.75±0.12  & 0.94±0.64 \\ \cline{2-4} 
\multicolumn{1}{|c|}{} & v-structure & 0.72±0.08  & 0.95±0.21 \\ \hline
\multicolumn{1}{|c|}{\multirow{3}{*}{\begin{tabular}[c]{@{}c@{}}7 variable \\ 3 cluster\end{tabular}}} & fork & 0.68±0.07 & 1.05±0.22 \\ \cline{2-4} 
\multicolumn{1}{|c|}{} & chain       & 0.60±0.06  & 0.97±0.5  \\ \cline{2-4} 
\multicolumn{1}{|c|}{} & v-structure & 0.65±0.06  & 1.0±0.0   \\ \hline
\multicolumn{2}{|c|}{Sachs}          & 0.51±0.051 & 0.57±0.49 \\ \hline
\multicolumn{2}{|c|}{Climate}        & 0.51±0.1   & 0.1±0.29  \\ \hline
\end{tabular}
\end{table}




\end{document}

%% file: math_commands.tex
\usepackage{amsmath,amsfonts,bm}

\def\1{\bm{1}}

\DeclareMathAlphabet{\mathsfit}{\encodingdefault}{\sfdefault}{m}{sl}
\SetMathAlphabet{\mathsfit}{bold}{\encodingdefault}{\sfdefault}{bx}{n}



%% file: main.bib
@inproceedings{niu2022learning,
  title     = {Learning Cluster Causal Diagrams: An Information-Theoretic Approach},
  author    = {Niu, Xueyan and Li, Xiaoyun and Li, Ping},
  booktitle = {Proceedings of the Thirty-First International Joint Conference on
               Artificial Intelligence, {IJCAI-22}},
  pages     = {4871--4877},
  year      = {2022},
  month     = {7},
}

@inproceedings{anand2023causal,
  title={Causal effect identification in cluster dags},
  author={Anand, Tara V and Ribeiro, Adele H and Tian, Jin and Bareinboim, Elias},
  booktitle={Proceedings of the AAAI Conference on Artificial Intelligence},
  volume={37},
  pages={12172--12179},
  year={2023}
}

@book{peters2017elements,
  title={Elements of causal inference: foundations and learning algorithms},
  author={Peters, Jonas and Janzing, Dominik and Sch{\"o}lkopf, Bernhard},
  year={2017},
  publisher={The MIT Press}
}

@inproceedings{annadani2021variational,
  author    = {Annadani, Yashas and Rothfuss, Jonas and Lacoste, Alexandre and Scherrer, Nino and Goyal, Anirudh and Bengio, Yoshua and Bauer, Stefan},
  title     = {Variational Causal Networks: Approximate Bayesian Inference over Causal Structures},
  booktitle = {BCIRWIS 2021: Workshop on Bayesian Causal Inference for Real World Interactive Systems (KDD 2021 Workshop)},
  year      = {2021}
}

@article{zheng2018dags,
  title={Dags with no tears: Continuous optimization for structure learning},
  author={Zheng, Xun and Aragam, Bryon and Ravikumar, Pradeep K and Xing, Eric P},
  journal={Advances in neural information processing systems},
  volume={31},
  year={2018}
}

@article{wahl2023foundations,
  author    = {Wahl, Jonas and Ninad, Urmi and Runge, Jakob},
  title     = {Foundations of causal discovery on groups of variables},
  journal   = {Journal of Causal Inference},
  volume    = {12},
  number    = {1},
  pages     = {20230041},
  year      = {2024},
  publisher = {De Gruyter},
  doi       = {10.1515/jci-2023-0041}
}

@article{semedo2020statistical,
  title={Statistical methods for dissecting interactions between brain areas},
  author={Semedo, Jo{\~a}o D and Gokcen, Evren and Machens, Christian K and Kohn, Adam and Byron, M Yu},
  journal={Current opinion in neurobiology},
  volume={65},
  pages={59--69},
  year={2020},
  publisher={Elsevier}
}

@article{perich2020rethinking,
  title={Rethinking brain-wide interactions through multi-region ‘network of networks’ models},
  author={Perich, Matthew G and Rajan, Kanaka},
  journal={Current opinion in neurobiology},
  volume={65},
  pages={146--151},
  year={2020},
  publisher={Elsevier}
}

@article{parviainen2017learning,
  title={Learning structures of Bayesian networks for variable groups},
  author={Parviainen, Pekka and Kaski, Samuel},
  journal={International Journal of Approximate Reasoning},
  volume={88},
  pages={110--127},
  year={2017},
  publisher={Elsevier}
}

@book{pearl2009causality,
  title={Causality},
  author={Pearl, Judea},
  year={2009},
  publisher={Cambridge university press}
}

@book{spirtes2000causation,
  title={Causation, prediction, and search},
  author={Spirtes, Peter and Glymour, Clark N and Scheines, Richard},
  year={2000},
  publisher={MIT press}
}

@inproceedings{rubenstein2017causal,
  author    = {Rubenstein, Paul K. and Weichwald, Sebastian and Bongers, Stephan and Mooij, Joris M. and Janzing, Dominik and Grosse-Wentrup, Moritz and Sch{\"o}lkopf, Bernhard},
  title     = {Causal Consistency of Structural Equation Models},
  booktitle = {Proceedings of the Thirty-Third Conference on Uncertainty in Artificial Intelligence (UAI 2017)},
  pages     = {808--817},
  year      = {2017}
}

@article{kawahara2010grouplingam,
  title={GroupLiNGAM: Linear non-Gaussian acyclic models for sets of variables},
  author={Kawahara, Yoshinobu and Bollen, Kenneth and Shimizu, Shohei and Washio, Takashi},
  journal={arXiv preprint arXiv:1006.5041},
  year={2010}
}

@inproceedings{wahl2023vector,
  title={Vector causal inference between two groups of variables},
  author={Wahl, Jonas and Ninad, Urmi and Runge, Jakob},
  booktitle={Proceedings of the AAAI Conference on Artificial Intelligence},
  volume={37},
  pages={12305--12312},
  year={2023}
}

@article{lorch2021dibs,
  title={Dibs: Differentiable bayesian structure learning},
  author={Lorch, Lars and Rothfuss, Jonas and Sch{\"o}lkopf, Bernhard and Krause, Andreas},
  journal={Advances in Neural Information Processing Systems},
  volume={34},
  pages={24111--24123},
  year={2021}
}

@inproceedings{geiger2013parameter,
  author    = {Geiger, Dan and Heckerman, David},
  title     = {Parameter Priors for Directed Acyclic Graphical Models and the Characterization of Several Probability Distributions},
  booktitle = {Proceedings of the Fifteenth Conference on Uncertainty in Artificial Intelligence},
  year      = {1999},
  pages     = {216--225},
  address   = {Stockholm, Sweden},
  publisher = {Morgan Kaufmann Publishers Inc.},
  series = {UAI'99}
}

@inproceedings{deleu2022bayesian,
  title={Bayesian structure learning with generative flow networks},
  author={Deleu, Tristan and G{\'o}is, Ant{\'o}nio and Emezue, Chris and Rankawat, Mansi and Lacoste-Julien, Simon and Bauer, Stefan and Bengio, Yoshua},
  booktitle={Uncertainty in Artificial Intelligence},
  pages={518--528},
  year={2022},
  organization={PMLR}
}

@article{hubert1985comparing,
  title={Comparing partitions},
  author={Hubert, Lawrence and Arabie, Phipps},
  journal={Journal of classification},
  volume={2},
  number={1},
  pages={193--218},
  year={1985},
  publisher={Springer}
}

@article{sachs2005causal,
  title={Causal protein-signaling networks derived from multiparameter single-cell data},
  author={Sachs, Karen and Perez, Omar and Pe'er, Dana and Lauffenburger, Douglas A and Nolan, Garry P},
  journal={Science},
  volume={308},
  number={5721},
  pages={523--529},
  year={2005},
  publisher={American Association for the Advancement of Science}
}

@article{blei2017variational,
  title={Variational inference: A review for statisticians},
  author={Blei, David M and Kucukelbir, Alp and McAuliffe, Jon D},
  journal={Journal of the American statistical Association},
  volume={112},
  number={518},
  pages={859--877},
  year={2017},
  publisher={Taylor \& Francis}
}

@article{williams1992simple,
  title={Simple statistical gradient-following algorithms for connectionist reinforcement learning},
  author={Williams, Ronald J},
  journal={Machine learning},
  volume={8},
  number={3},
  pages={229--256},
  year={1992},
  publisher={Springer}
}

@inproceedings{eggeling2019structure,
  title={On structure priors for learning Bayesian networks},
  author={Eggeling, Ralf and Viinikka, Jussi and Vuoksenmaa, Aleksis and Koivisto, Mikko},
  booktitle={The 22nd International Conference on Artificial Intelligence and Statistics},
  pages={1687--1695},
  year={2019},
  organization={PMLR}
}

@inproceedings{DAGBootstrap,
author = {Friedman, Nir and Goldszmidt, Moises and Wyner, Abraham},
title = {Data analysis with bayesian networks: a bootstrap approach},
year = {1999},
isbn = {1558606149},
publisher = {Morgan Kaufmann Publishers Inc.},
address = {San Francisco, CA, USA},
booktitle = {Proceedings of the Fifteenth Conference on Uncertainty in Artificial Intelligence},
pages = {196–205},
numpages = {10},
location = {Stockholm, Sweden},
series = {UAI'99}
}

@inproceedings{chalupka2016multi,
  title={Multi-level cause-effect systems},
  author={Chalupka, Krzysztof and Eberhardt, Frederick and Perona, Pietro},
  booktitle={Artificial intelligence and statistics},
  pages={361--369},
  year={2016},
  organization={PMLR}
}

@article{lievin2020optimal,
  title={Optimal variance control of the score-function gradient estimator for importance-weighted bounds},
  author={Li{\'e}vin, Valentin and Dittadi, Andrea and Christensen, Anders and Winther, Ole},
  journal={Advances in Neural Information Processing Systems},
  volume={33},
  pages={16591--16602},
  year={2020}
}

@inproceedings{zheng2020learning,
  title={Learning sparse nonparametric dags},
  author={Zheng, Xun and Dan, Chen and Aragam, Bryon and Ravikumar, Pradeep and Xing, Eric},
  booktitle={International conference on artificial intelligence and statistics},
  pages={3414--3425},
  year={2020},
  organization={Pmlr}
}
